\documentclass[letterpaper]{article}

\usepackage{microtype}
\usepackage{graphicx}
\usepackage{float}
\usepackage{stfloats}
\usepackage{placeins}
\usepackage[center]{caption}
\usepackage{subcaption}
\usepackage{booktabs}

\usepackage{hyperref}

\newcommand{\demopageurl}{https://google.github.io/tacotron/publications/text_predicting_global_style_tokens/index.html}

\usepackage{slt2018}

\title{Predicting Expressive Speaking Style from Text in End-to-End Speech Synthesis}

\usepackage{todonotes}

\begin{document}

\name{Daisy Stanton, Yuxuan Wang, RJ Skerry-Ryan}
\address{Google, Inc. \\
         1600 Amphitheatre Parkway \\
         Mountain View, CA 94043}

\maketitle

\begin{abstract}
Global Style Tokens (GSTs) are a recently-proposed method to learn latent disentangled representations of high-dimensional data. GSTs can be used within Tacotron, a state-of-the-art end-to-end text-to-speech synthesis system, to uncover expressive factors of variation in speaking style. In this work, we introduce the Text-Predicted Global Style Token (TP-GST) architecture, which treats GST combination weights or style embeddings as ``virtual'' speaking style labels within Tacotron. TP-GST learns to predict stylistic renderings from text alone, requiring neither explicit labels during training, nor auxiliary inputs for inference. We show that, when trained on a dataset of expressive speech, our system generates audio with more pitch and energy variation than two state-of-the-art baseline models. We further demonstrate that TP-GSTs can synthesize speech with background noise removed, and corroborate these analyses with positive results on human-rated listener preference audiobook tasks. Finally, we demonstrate that multi-speaker TP-GST models successfully factorize speaker identity and speaking style. We provide a website with audio samples \footnote{\url{https://google.github.io/tacotron/publications/text_predicting_global_style_tokens}} for each of our findings.
\end{abstract}


\section{Introduction}
\label{sec.intro}
A major challenge for modern text-to-speech (TTS) research is developing models that can produce a natural-sounding speaking style for a given piece of text input.
Part of the challenge is that many factors contribute to ``natural-sounding'' speech,
including high audio fidelity, correct pronunciation, and what is known as good \textit{prosody}. 
Prosody includes low-level characteristics such as pitch, stress, breaks, and rhythm, and impacts speaking \textit{style}, which describes higher-level characteristics such as emotional valence and arousal. 
Prosody and style are particularly difficult to model, as they encompass information typically not specified in text: there are many different~\textendash~yet valid~\textendash~renderings of the same piece of text.  Additionally, while considerable effort has been spent modeling such renderings using annotations, explicit labels are difficult to define precisely, costly to acquire, noisy in nature, and don't necessarily correlate with perceptual quality. 

Tacotron \cite{yx2017tacotron} is a state-of-the-art speech synthesis system that computes its output directly from graphemes or phonemes. Like many modern TTS systems, it learns an implicit model of prosody from statistics of the training data alone.  It can learn, for example, to inflect English phrases ending in a question mark with a rise in pitch. As noted in \cite{wang2018gst}, however, synthesizing long-form expressive datasets (such as audiobooks) presents a challenge, since wide-ranging voice characteristics are collapsed into a single, ``averaged'' model of prosodic style. 
While \cite{wang2018gst} learns disentangled factors of speaking style within Tacotron, it requires either audio or manually-selected weights at inference time to generate output. 
Given all of the above, of particular interest would be a speech synthesis system that not only learns to represent a wide range of speaking styles, but that can synthesize expressive speech without the need for auxiliary inputs at inference time. In this work, we aim to do just that. 
Our main contribution is a pair of extensions to Global Style Tokens (GSTs) \cite{wang2018gst} that predict speaking style from text. The two alternative prediction pathways are easy to implement and require no additional labels.
We show that, like baseline GST models, our system can capture speaker-independent factors of variation, including speaking style and background noise. We provide audio samples, analysis, and results showing that our models are significantly preferred in subjective evaluations.

\section{Related Work}

Attempts to model prosody and speaking style span more than three decades in the statistical speech synthesis literature. These methods, however, have largely required explicit annotations, which pose the difficulties discussed in Section \ref{sec.intro}. INTSINT \cite{hirst1987intsint}, ToBi \cite{silverman1992tobi}, Momel \cite{hirst1993momel}, landmark detection \cite{liu1996landmark}, Tilt \cite{taylor1998tilt}, and SLAM \cite{obin2014slam} all describe methods to annotate or classify prosodic features such as breaks, intonation, rhythm, and melody.  Notable among these is AuToBI \cite{rosenberg2010autobi}, which automatically detects and classifies these features, but which requires models pretrained on labeled data to do so. 

Substantial effort has also gone into modeling emotion, but these methods, too, have traditionally required keywords, semantic representations, or labels for model training. 
Recent examples include
\cite{lee2015RNN_Emotion}, \cite{wang2017UtteranceLevelEmotion}, \cite{khorram2017CNNs_For_Emotion}, \cite{latif2017vae_supervised_emotion}, and \cite{lorenzo2018investigating}, \cite{deng2018semisupervised_emotion}.

\cite{jauk2017unsupervised} explores various methods to predict acoustic features such as $i$-vectors \cite{dehak2011front} from semantic embeddings. These methods rely on a complex set of hand-designed features, however, and require training three models in separate steps (the acoustic feature predictor, a neutral-prosody synthesis model, and a speaker-adaptation model).

The recently-published VAE-Loop \cite{akuzawa2018vaeloop} aims to learn speaking style variations by conditioning VoiceLoop \cite{taigman2017voiceloop}, an autoregressive speech synthesis model, on the global latent variable output by a conditional variational autoencoder (VAE).
%While this work claims to enable unsupervised control of synthesized speech expressiveness, 
In inference mode, however, the latent variable $z$ still needs to be fed into the model to achieve control. Furthermore, while 
$z$ is expected to acquire latent representations of global speaking styles, the experimental analysis (\cite{akuzawa2018vaeloop}, section 4.5) and audio samples \cite{akuzawa2018vaeloop_demopage} suggest that $z$ has primarily learned speaker gender and identity rather than prosody or speaking style. 
\begin{figure*}[p]
\centering
\begin{subfigure}{0.76\textwidth}
    \includegraphics[scale=0.76]{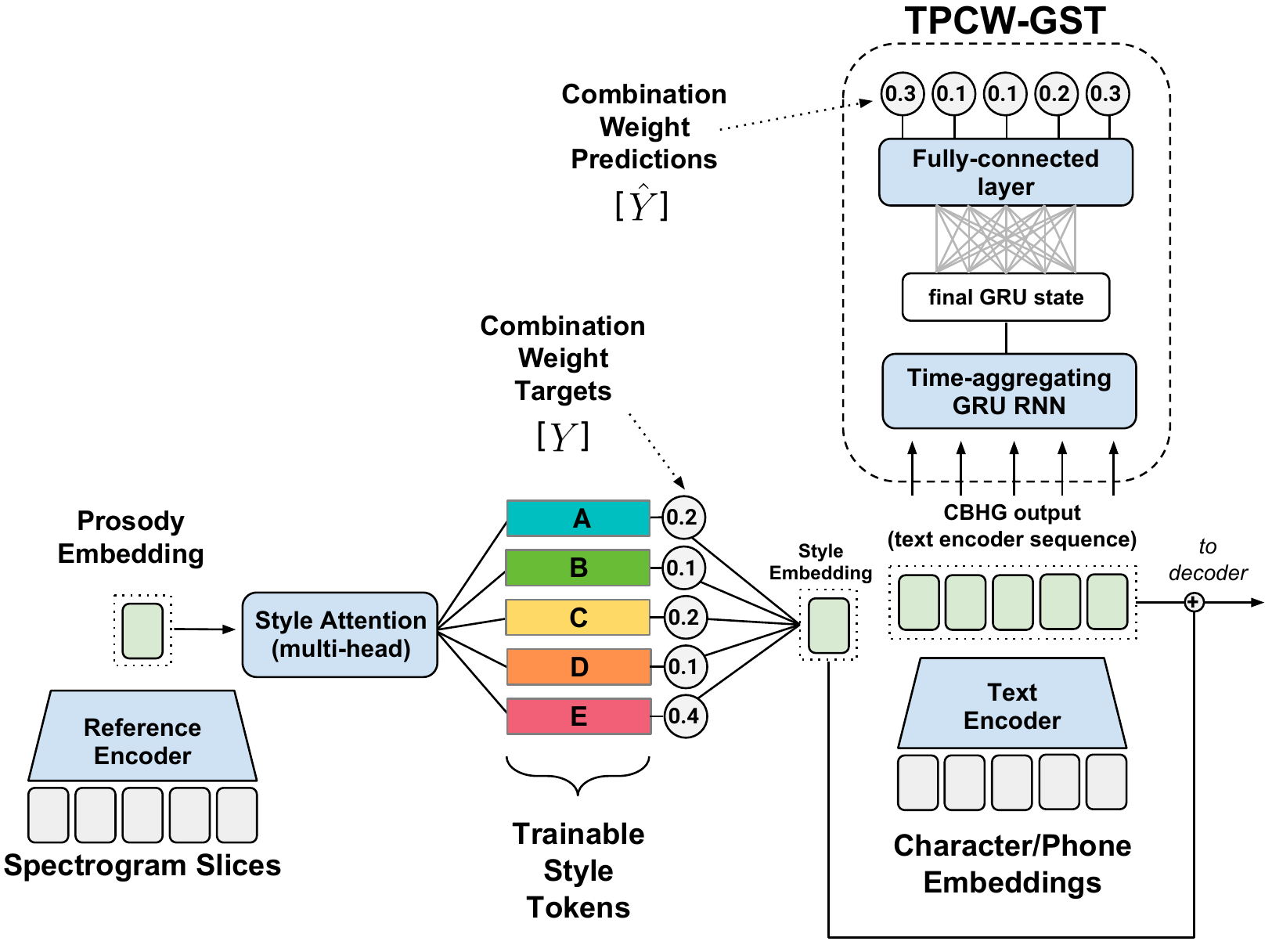}
    \vspace*{0.4cm}    \captionsetup{margin={-1.5cm,-1.5cm}}
    \caption{TPCW-GST architecture, the first of two possible prediction pathways. This pathway uses the GST combination weights as targets during training, and adds an additional cross-entropy$(Y, \hat{Y})$ term to the Tacotron loss function.\label{fig.diagram_textpredict_comb_weights}}    
\end{subfigure}
\\
\vspace*{1.0cm}\begin{subfigure}{0.76\textwidth}
    \includegraphics[scale=0.76]{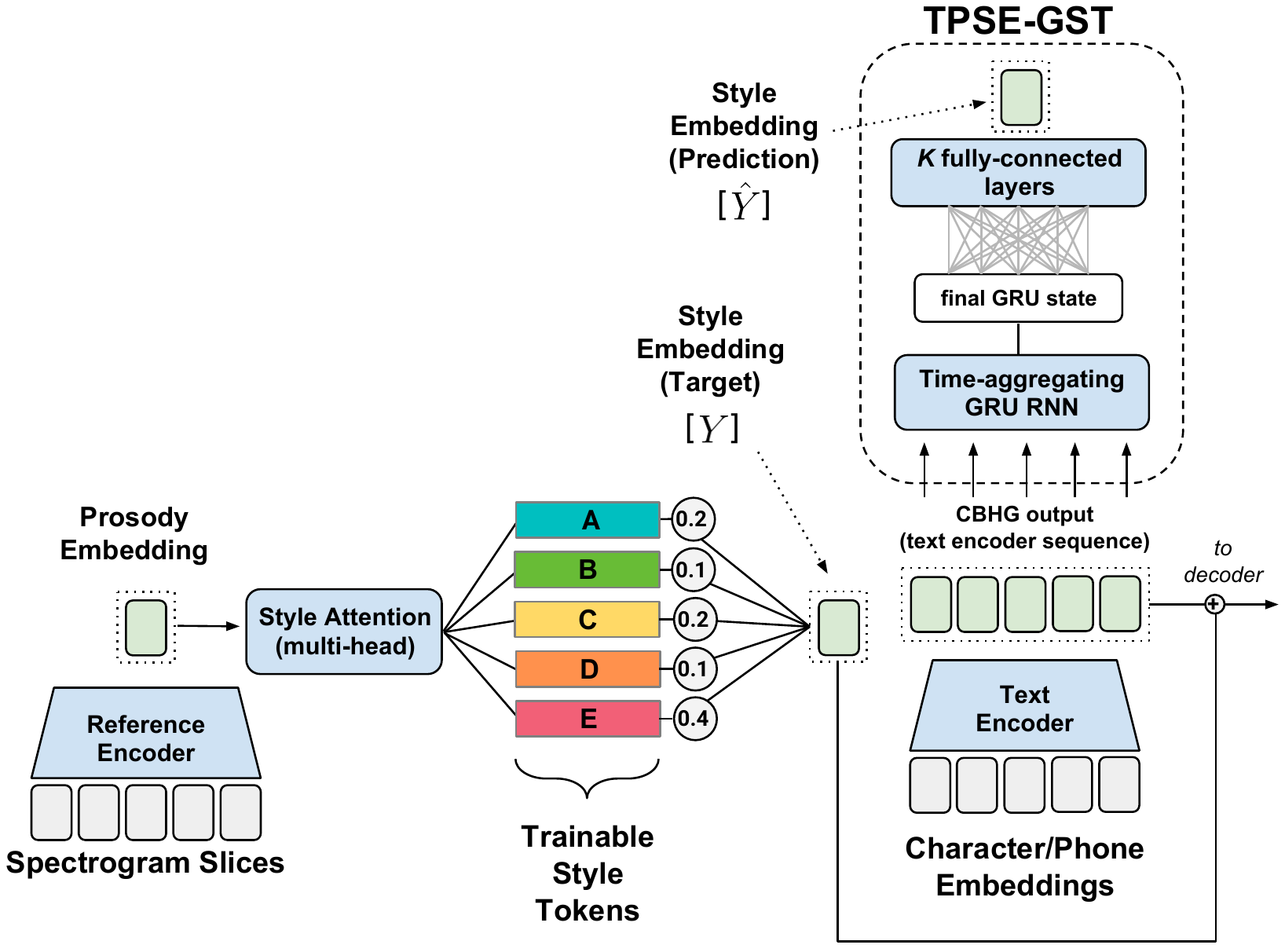}
    \vspace*{0.4cm}    \captionsetup{margin={-1.5cm,-1.5cm}}
    \caption{TPSE-GST architecture, the second of two possible prediction pathways. This pathway treats the GST style embedding as a target during training, and adds an additional $L_1(Y, \hat{Y})$ term to the Tacotron loss function.\label{fig.diagram_textpredict_style_embedding}}    
\end{subfigure}
\caption{TP-GST architectures.\label{fig.model.diagrams}}
\end{figure*}

\vspace*{0.3cm}
\section{Model}
\label{sec.model}

Our model is based on an augmented version of Tacotron \cite{yx2017tacotron}, a recently proposed state-of-the-art speech synthesis model that predicts mel spectrograms directly from grapheme or phoneme sequences. 

The augmented version we use is the Global Style Token (GST) \cite{wang2018gst} architecture, which adds to Tacotron
a spectrogram reference encoder \cite{rj2018transfer}, a style attention module, and a style embedding for conditioning. During training, the \textbf{style attention} learns to represent the reference encoder output (called the \textbf{prosody embedding}) as a convex combination of trainable embeddings called \textit{style tokens}. These are shared across all utterances in the training set, and capture global variation in the data -- hence the name Global Style Tokens. We call the convex combination of style tokens the \textbf{style embedding}.

Our proposed architecture, which we call ``Text-Predicted Global Style Tokens'' (TP-GST), adds two possible text-prediction pathways to a GST-enhanced Tacotron. These allow the system to predict style embeddings at inference time by either:
%\vspace{-0.1cm}
\begin{enumerate}
    \item interpolating the GSTs learned during training, using \textbf{combination weights} predicted only from the text (``TPCW''); or
    %\vspace{-0.05cm}
    \item directly predicting \textbf{style embeddings} from text features, ignoring style tokens and combination weights (``TPSE'').
\end{enumerate}
%\vspace{-0.1cm}
\pagebreak

Using operators to stop gradient flow, the two text-prediction pathways can be trained jointly. 
At inference time, the model can be run as a TPCW-GST, as a TPSE-GST, or (by supplying auxiliary inputs) as a traditional GST-Tacotron. 

We describe each of the two text prediction pathways in more detail below.

\vspace*{0.3cm}
\subsection{Text features}

Both TP-GST pathways use as features the output of Tacotron's text encoder. This output is computed by an encoder sub-module called a \textbf{CBHG} \cite{yx2017tacotron}, which explicitly models local and contextual information in the input sequence. A CBHG consists of a bank of 1-D convolutional filters, followed by highway networks \cite{srivastava2015highway} and a bidirectional Gated Recurrent Unit (GRU) \cite{cho2014gru} recurrent neural net (RNN).  Since the text encoder outputs a variable-length
sequence, the first step of TP-GST is to pass this sequence through a 64-unit time-aggregating GRU-RNN, and use its final output as a fixed-length text feature vector. The GRU-RNN acts as a summarizer for the text encoder in much the same way as the 128-unit GRU-RNN in \cite{rj2018transfer} acts as a summarizer for the reference encoder; both time-aggregate their variable-length input and output a fixed-length summary.

The fixed-length text features are used as input in both text prediction pathways, discussed next. 

\vspace*{0.3cm}
\subsection{Predicting Combination Weights (TPCW)}

In the GST-augmented Tacotron, a reference signal's prosody embedding serves as the \textit{query} to an attention mechanism over the style tokens, and the resulting \textit{values}, normalized via a softmax activation, serve as the combination weights.
As illustrated in Figure \ref{fig.diagram_textpredict_comb_weights}, the simpler version of our model treats these GST combination weights as a prediction target during training. We call this system TPCW-GST, to stand for ``text-predicted combination weights''. Note that since the style attention and style tokens are updated via backpropagation, the GST combination weights form moving targets during training.

To learn to predict these weights, we feed the fixed-length text features from the time-aggregating GRU-RNN to a fully-connected layer. 
The outputs of this layer are treated as logits, and we compute the cross-entropy loss between these values and the (target) combination weights output by the style attention module. We stop the gradient flow to ensure that text prediction error doesn't backpropagate through the GST layer, and add the cross-entropy result to the final Tacotron loss. 

At inference time, the style tokens are fixed, and this pathway can be used to predict the token combination weights from text features alone.

\subsection{Predicting Style Embeddings (TPSE)}

Figure \ref{fig.diagram_textpredict_style_embedding} illustrates 
a second, alternative, prediction pathway. We call this system TPSE-GST, to stand for ``text-predicted style embeddings''.  This version of the model feeds the text feature sequence through one or more fully-connected layers, and outputs a style embedding prediction directly. We train this pathway using an $L_1$ loss between the predicted (TPSE-GST) and target (GST) style embeddings. As is done for a TPCW-GST, we stop the gradient flow to ensure that text prediction error doesn't backpropagate through the GST layer.

\begin{figure*}[bt]
\centering
\vspace{-0.2cm}
\begin{subfigure}{0.95\textwidth}
    \includegraphics[scale=0.54]{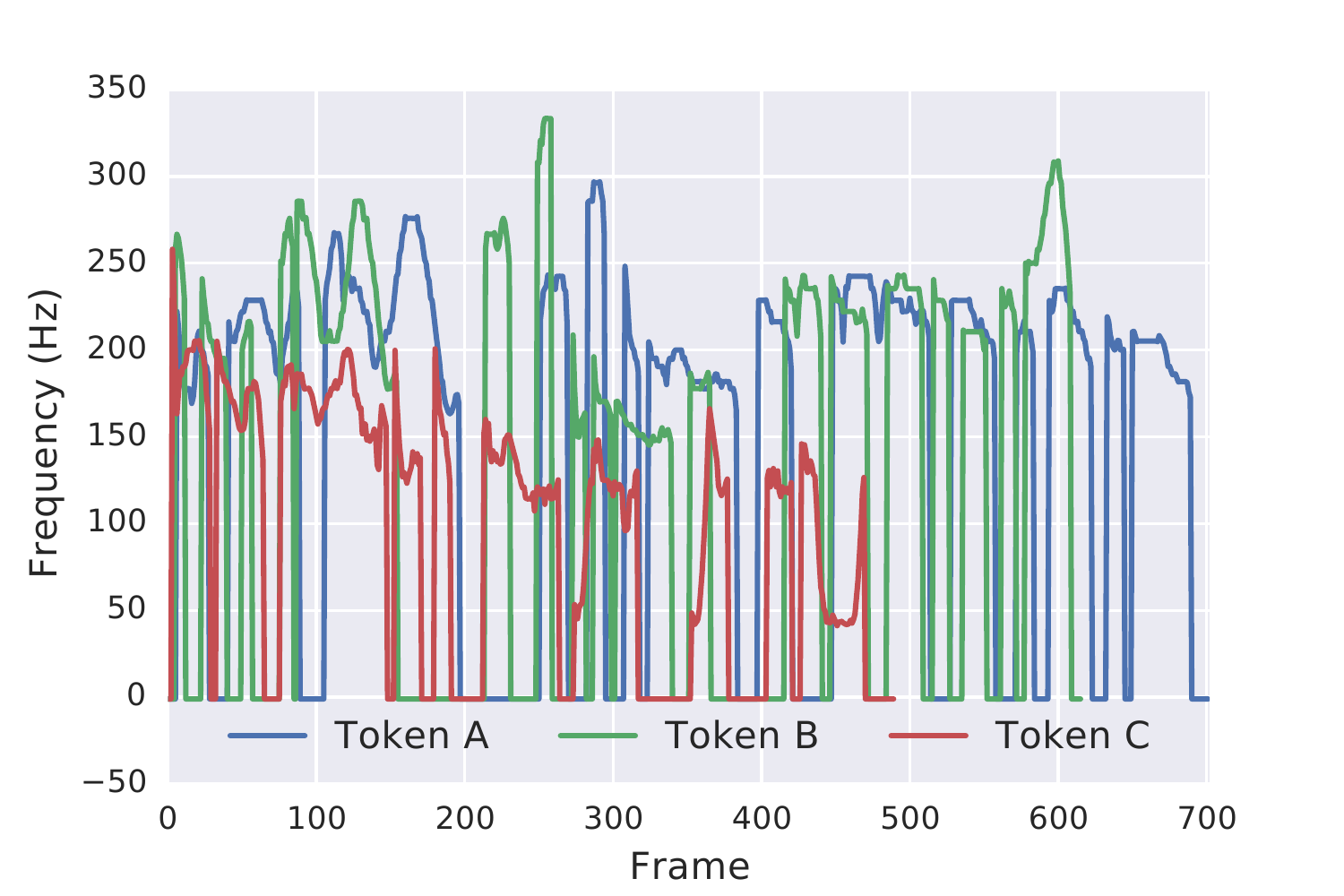}
    \includegraphics[scale=0.54]{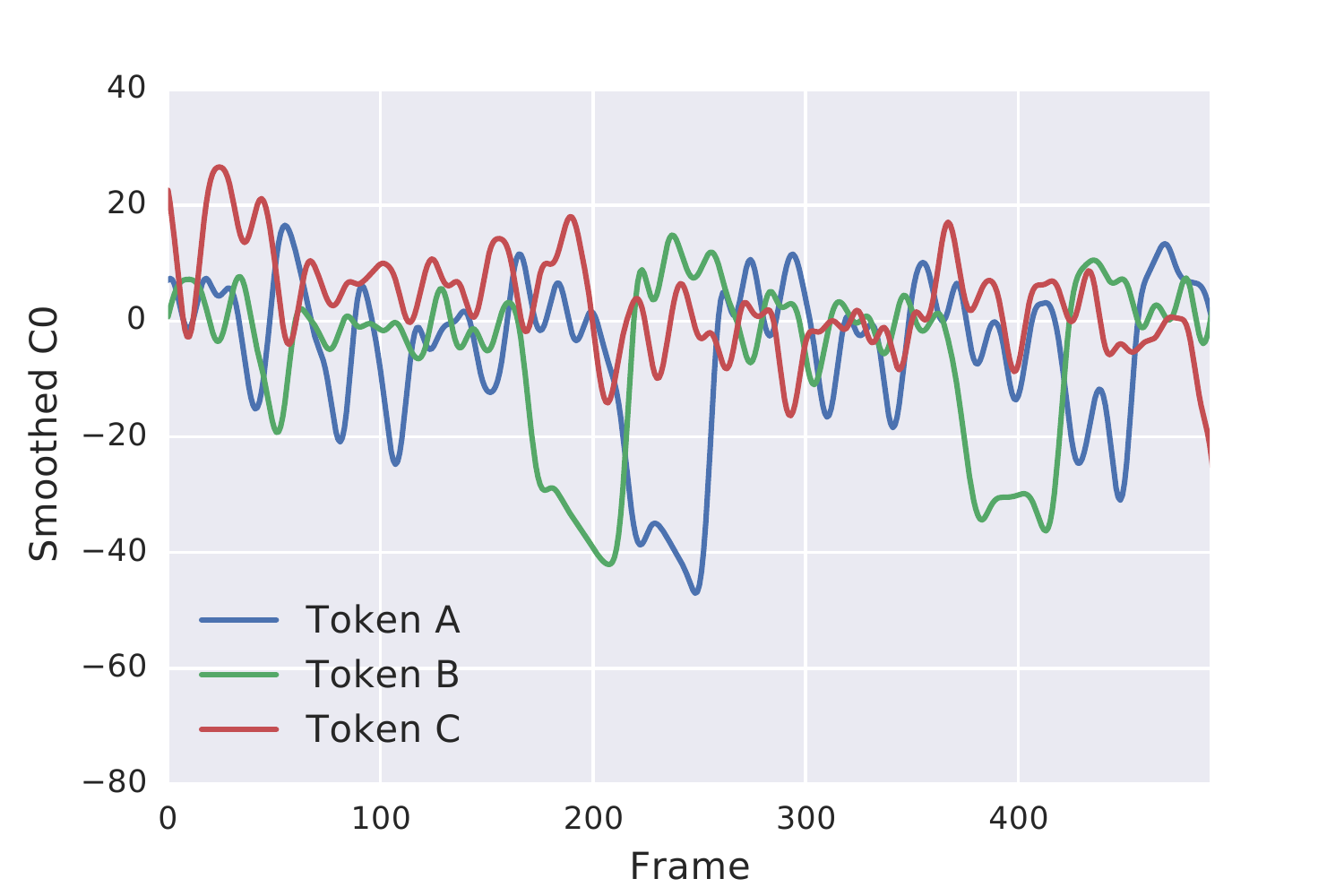}
    \vspace{-0.17cm}
    \caption{$F_0$ and log-$C_0$ for three tokens}
    \vspace{-0.05cm}   
\end{subfigure}
\begin{subfigure}{1.0\textwidth}
    \hspace{-0.6cm}
    \includegraphics[scale=0.452]{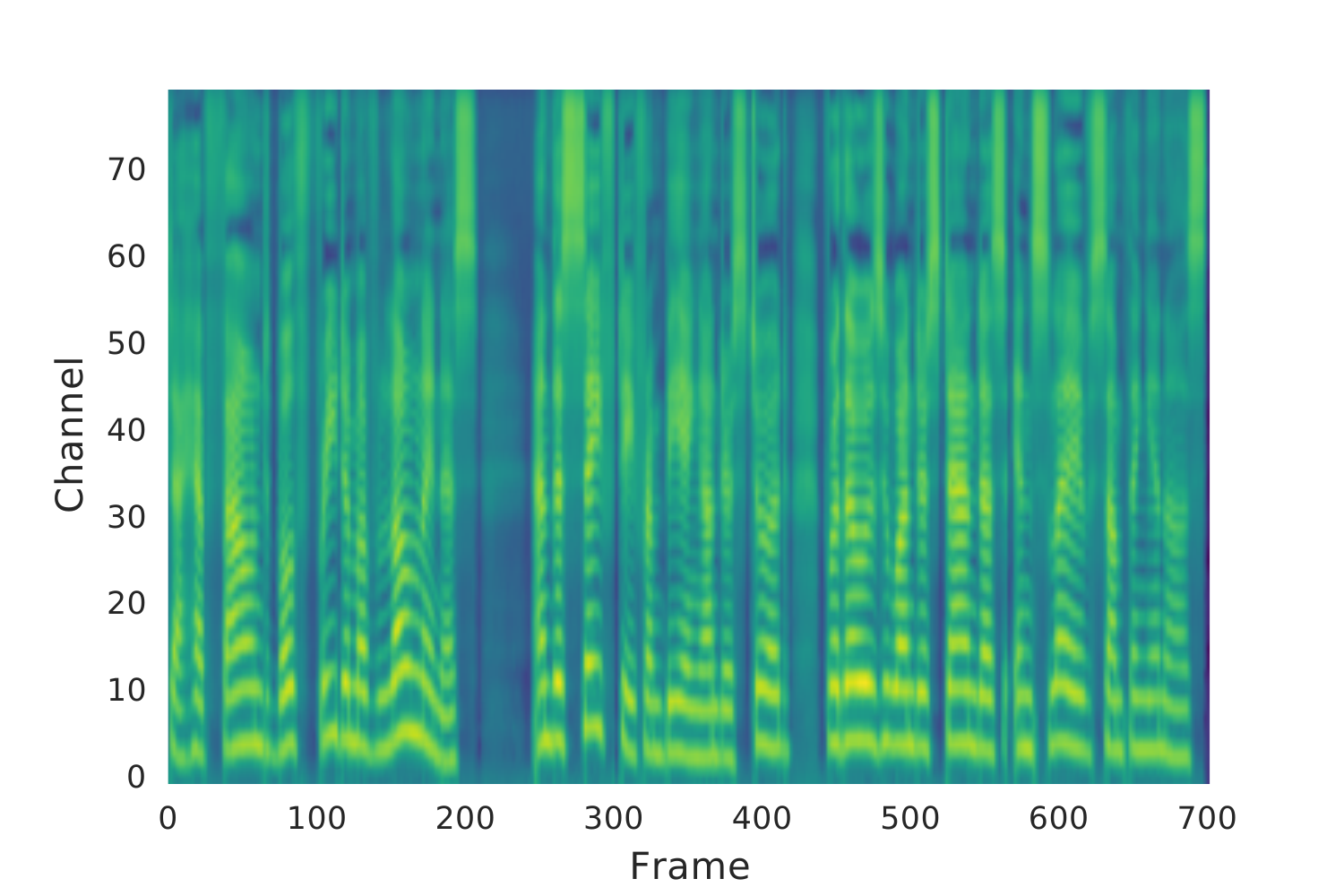}
    \hspace{-1.05cm}
    \includegraphics[scale=0.452]{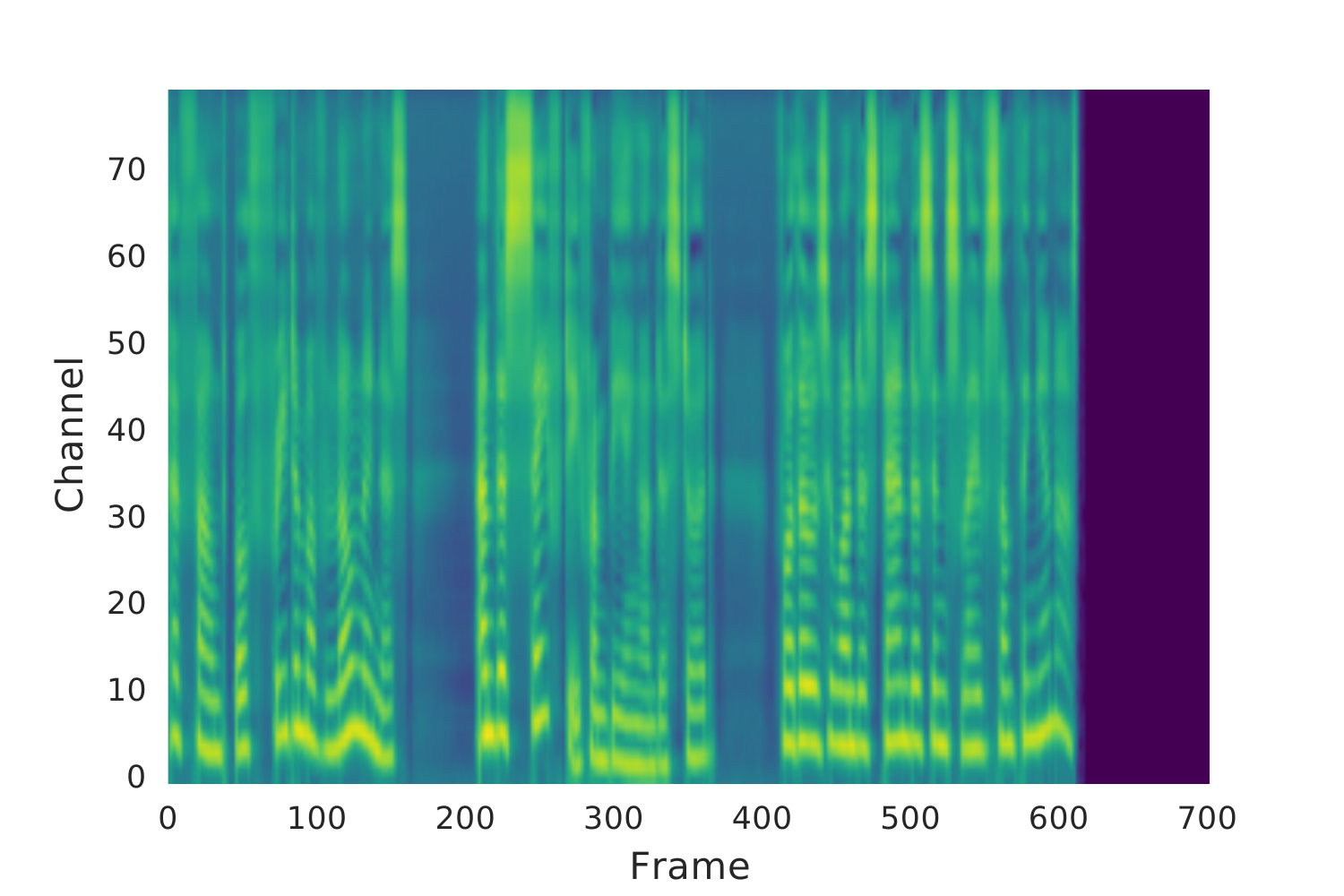}
    \hspace{-1.05cm}
    \includegraphics[scale=0.452]{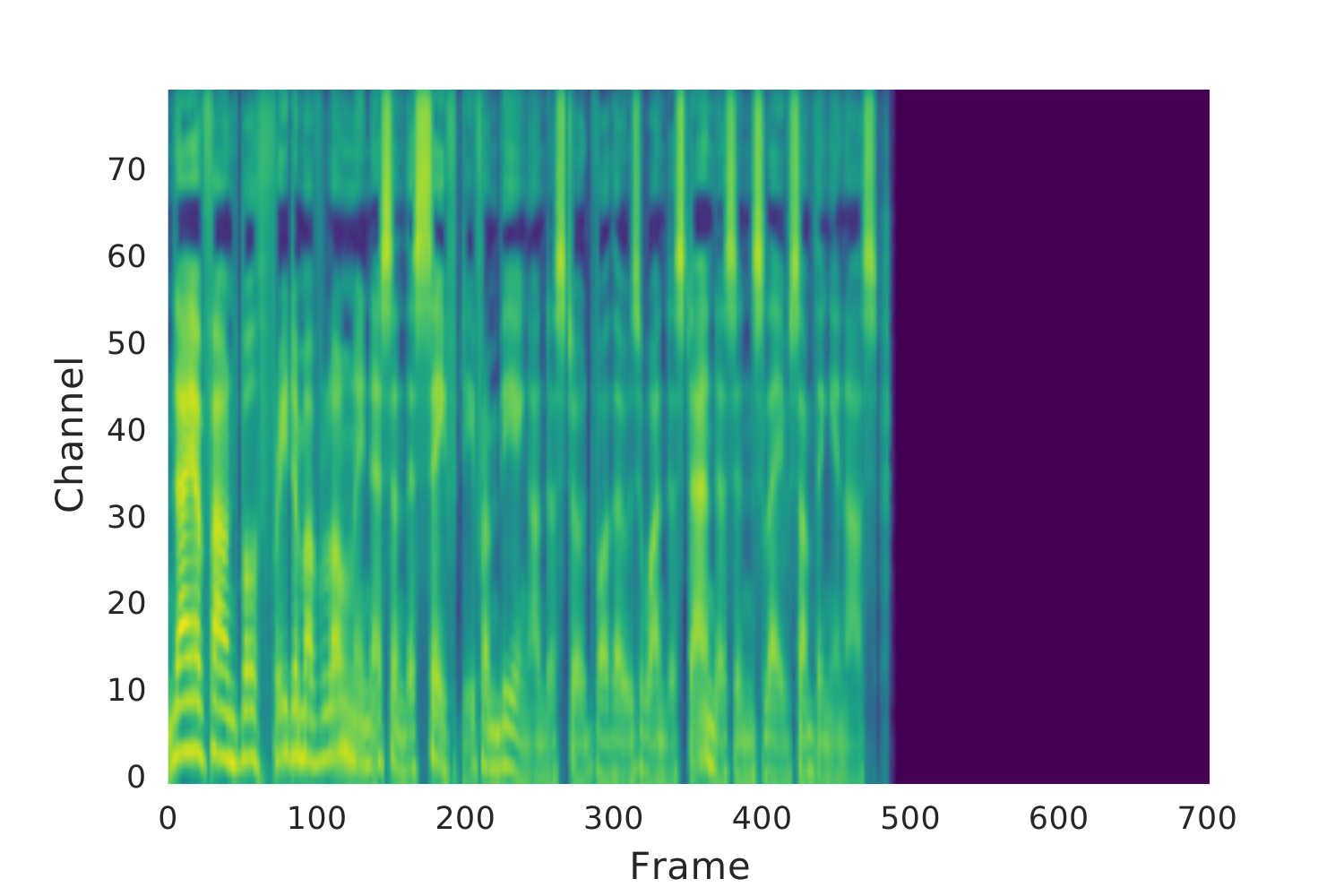}
    \captionsetup{margin={0cm,0.6cm}}
    \caption{Mel spectrograms for the three tokens above}    
    \vspace{0.3cm}
\end{subfigure}
\caption{(a) $F_0$ and log-$C_0$ of an audiobook phrase, synthesized using three tokens from a single-speaker TP-GST Tacotron.\\*
(b) Mel-scale spectrograms of the same phrase corresponding to each token. See text for details.\label{fig.lessac.tokens}}
\par
\par
\end{figure*}

%\vspace{0.35cm}
We use ReLU activations for the hidden fully-connected layers, and a tanh activation on the output layer that emits the text-predicted style embedding. This is intended to match the style token tanh activation (see \cite{wang2018gst}, section 3.2.2), which, in turn, is chosen to match the GRU tanh activation of the final bidirectional RNN in the text encoder CBHG \cite{yx2017tacotron}. As in \cite{wang2018gst}, this choice leads to better token variation.

In inference mode, this pathway can be used to predict the style embedding directly from text features. Note that the model completely ignores the style tokens in this mode, since they are not needed: they are only used to compute the style embedding prediction target during training.

\begin{figure*}[bt]
\centering
\begin{subfigure}{0.9\textwidth}
    \includegraphics[scale=0.495]{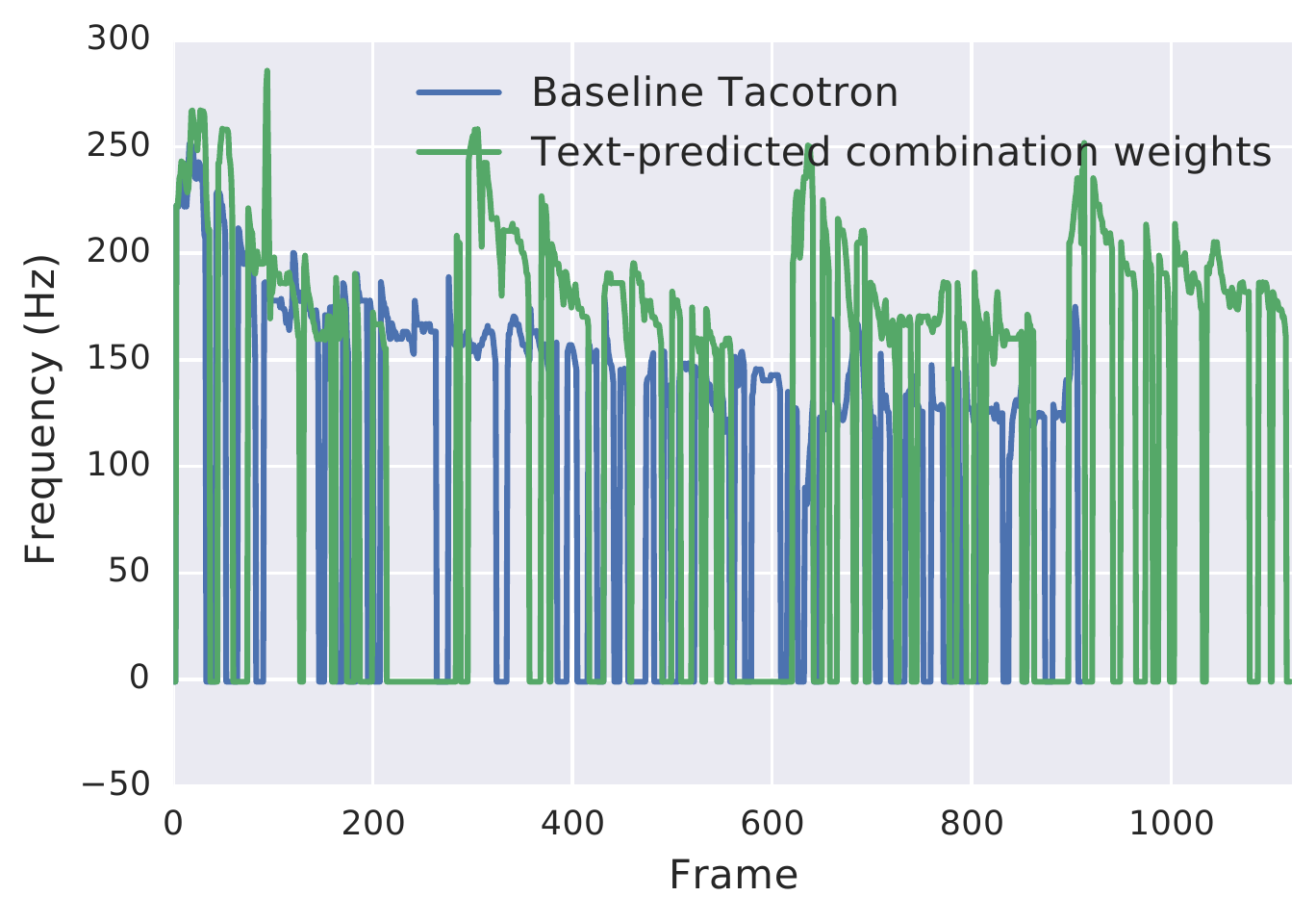}
    \hspace{0.3cm}
    \includegraphics[scale=0.495]{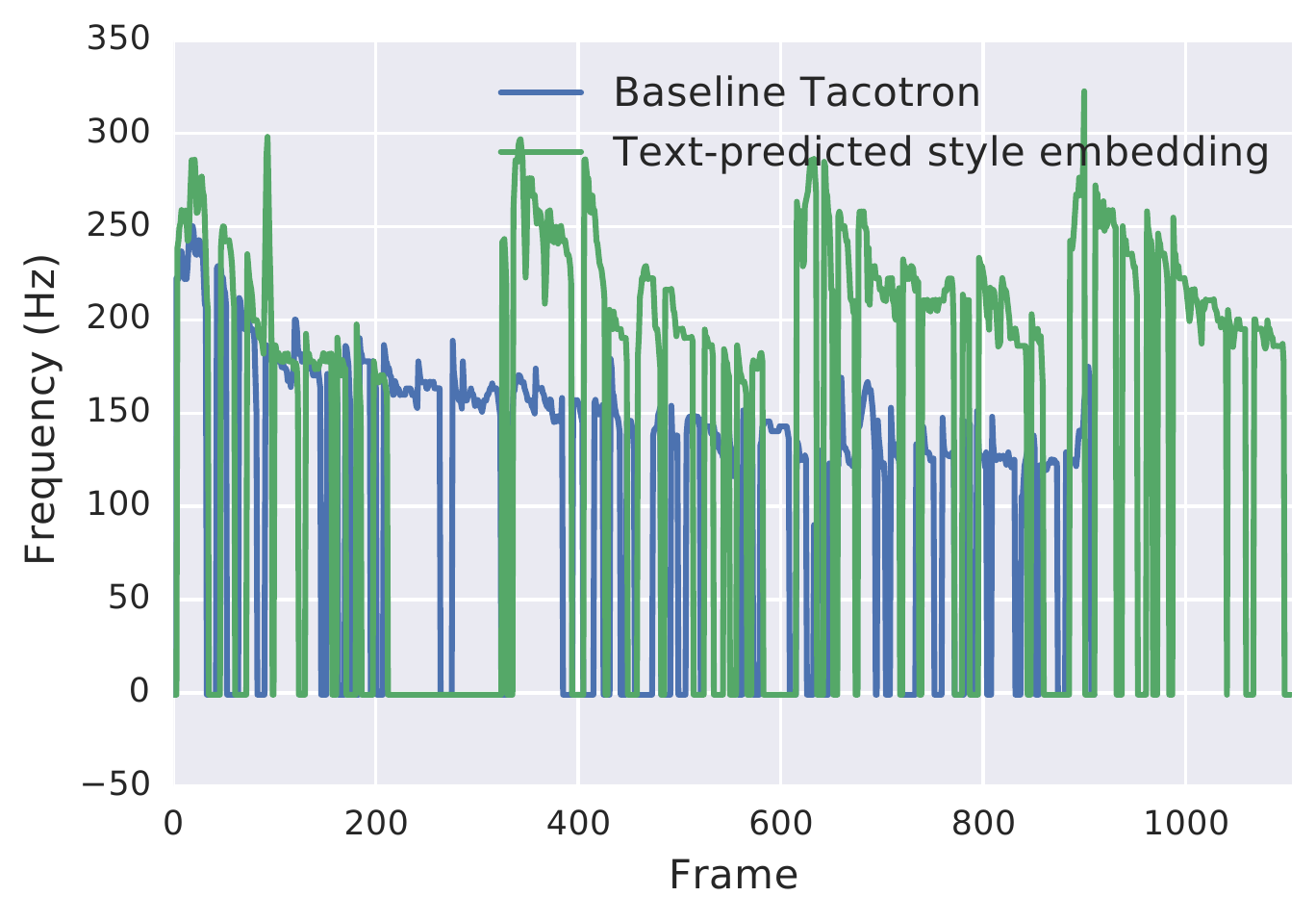}
    \vspace{-0.15cm}
    \caption{Phrase $F_0$ from both TP-GST systems vs Tacotron}    
    \vspace*{0.4cm}
\end{subfigure}
\begin{subfigure}{0.9\textwidth}
    \hspace{0.3cm}
    \includegraphics[scale=0.477]{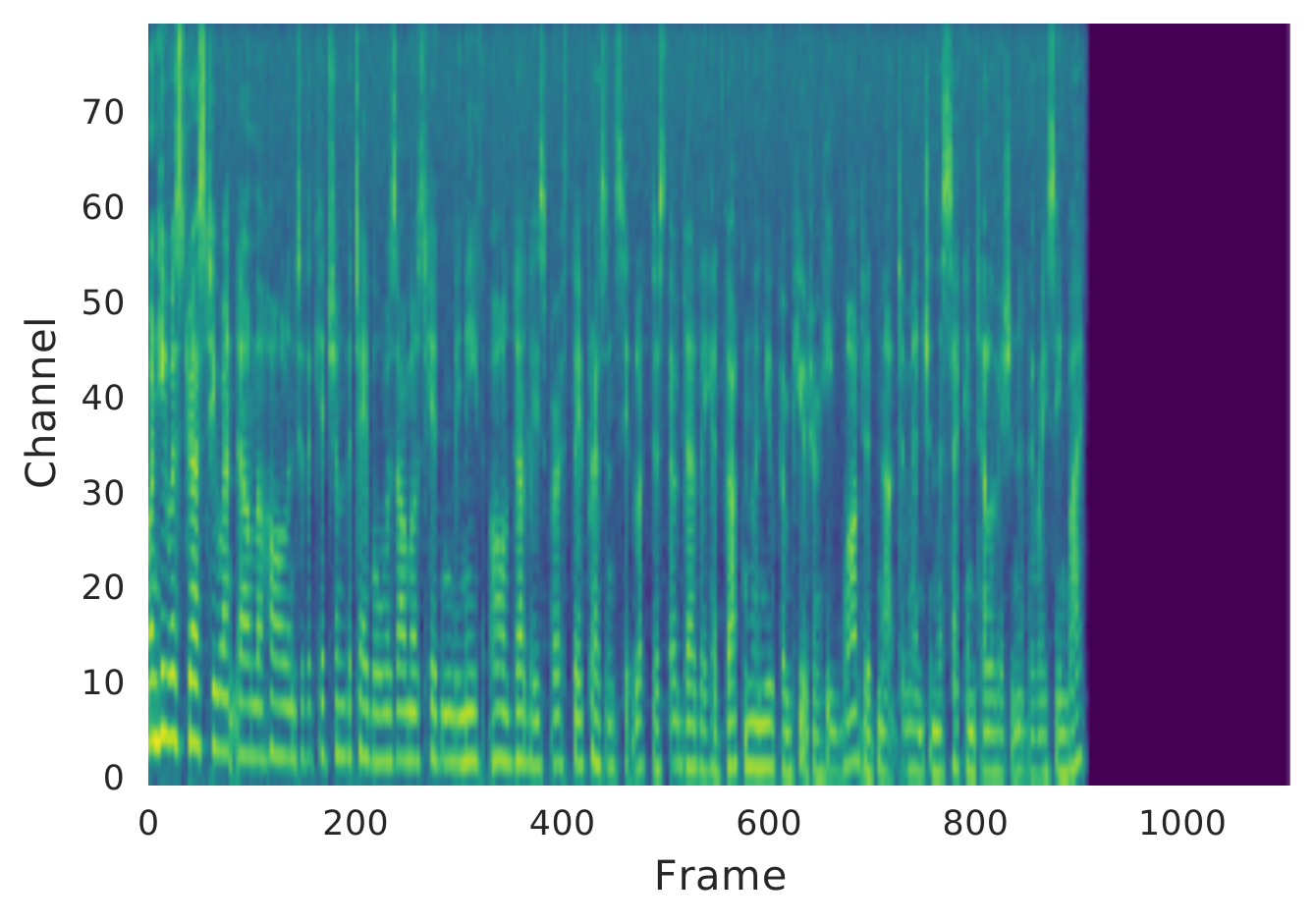}
    \hspace{0.6cm}
    \includegraphics[scale=0.477]{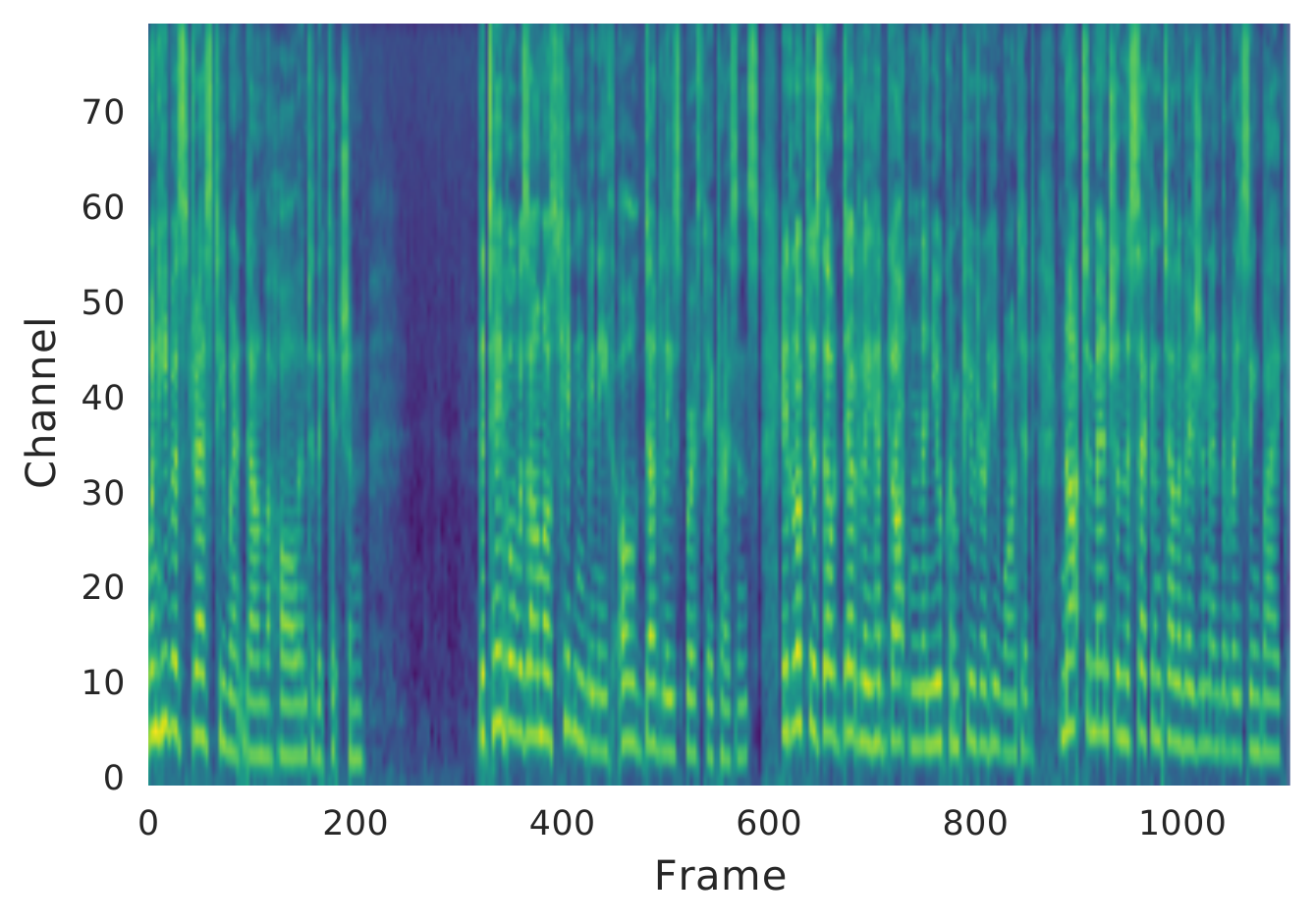}
    \vspace{-0.02cm}
    \caption{Phrase mel spectrograms for baseline Tacotron (left) and TPSE-GST (right)}
    \vspace{0.3cm}
\end{subfigure}
\caption{The same phrase, unseen during training, synthesized using a baseline Tacotron, TPCW-GST, and TPSE-GST. Notice that the baseline's ``declining pitch'' problem is fixed by both text-prediction systems.\label{fig.lessac.phrases}}
\end{figure*}

\section{Experiments}
\label{sec.expt}

In this section, we evaluate the performance of synthesis using TP-GST; we examine both single- and multi-speaker models. As is common for generative models, objective metrics often do not correlate well with perception \cite{theis2015note}. While we use visualizations for some experiments below, we strongly encourage readers to listen to the audio samples provided on our demo page.

\subsection{Single Speaker Experiments}
\label{sec.expt.singlespeaker}

Our single-speaker TP-GST model is trained on 147 hours of American English audiobook data. The books are read by the 2013 Blizzard Challenge speaker, Catherine Byers, in an animated and emotive storytelling style. Some books contain very expressive character voices with high dynamic range, which are challenging to model. 

The model uses 20 Global Style Tokens with 4-headed additive attention, and predicts both TP-GST targets (TPCW and TPSE) during training. While the number of hidden layers in the TPSE-GST pathway is configurable (see Figure \ref{fig.diagram_textpredict_style_embedding}), these experiments use a single hidden layer of size $64$. We train all models with a minibatch size of 32 using the Adam optimizer \cite{kingma2014adam}, and perform evaluations at about 250,000 steps.

\label{sec.expt.singlespeaker.style_token_variation}
\subsubsection{Style Token Variation} 
Token variation is key for a GST model to be able to represent a large range of expressive styles in speech data. As in a GST-Tacotron, we can verify that a TP-GST system learns a rich set of style tokens during training. Figure \ref{fig.lessac.tokens} shows the fundamental frequency ($F_0$) and smoothed average power (log-$C_0$) of three different style tokens learned from this system, superimposed to highlight their variation. It also shows spectrograms generated by conditioning the model on each token in turn. These plots visualize what can be heard on our samples page, which is that different tokens capture variation in pitch, energy, speaking rate. As in a standard GST-Tacotron, conditioning the model on a particular token will result in the same $F_0$ and $C_0$ trend relative to the others, independent of the text input.

\subsubsection{Synthesis with Style Prediction}
In addition to verifying token variation,
we can compare synthesis output to that of a vanilla Tacotron model. This allows us to contrast what two Tacotron-based systems can predict from text alone.
Note that we choose the vanilla Tacotron as a baseline, since comparing TP-GST to a GST-Tacotron is not apples to apples: a GST-Tacotron requires either a reference signal or a manual selection of style token weights at inference time.

Figure \ref{fig.lessac.phrases} shows $F_0$ contours and mel spectrograms generated by a baseline Tacotron model and both pathways of TP-GST model (20 tokens, 4 heads). 
Each depicts the same audiobook phrase unseen during training. We see that the TP-GST model yields a more varied $F_0$ contour and richer spectral detail. This example also highlights a point noted in \cite{yx2017tacotron}, which is that baseline Tacotron models trained on expressive speech can result in synthesis with a continuously declining pitch (green curve). Like a GST Tacotron, we see that the TP-GST model fixes this problem, but without needing a reference signal for inference.

\subsubsection{Subjective Evaluation}
\label{sec.expt.singlespeaker.sxs}
To evaluate the quality of this method at scale, we provide side-by-side subjective test results of TP-GST synthesis versus a baseline Tacotron. The evaluation data used for this test was a set of 260 sentences from an audiobook unseen during training, including many long phrases. In each test, raters listened to the same sentence synthesized by both a baseline Tacotron and one of the TP-GST systems. They then evaluated the pair on a 7-point Likert scale ranging from ``much worse than'' to ``much better than''; each comparison received 8 scores from different raters.

The results are shown in Table \ref{table.lessac_sxs_tp_vs_tacotron}. In both subjective tests, raters preferred the text-predicted style enhancements over the Tacotron baseline.

\vspace*{0.4cm}
\begin{table}[h]
\begin{center}
\begin{small}
\begin{sc}
\setlength\tabcolsep{3.5pt} \begin{tabular}{c|ccc|cc}
\toprule
& \multicolumn{3}{c|}{preference (\%)} & \multicolumn{2}{c} {p-value}
\\ \cline{2-6}
& baseline & neutral & TP-GST & 3-point & 7-point \\
\midrule
TPCW & 18.6\% &22.6\% &58.8\% &$< 10^{-37}$ & $< 10^{-35}$ \\
TPSE & 17.5\% &24.7\% &57.8\% &$< 10^{-38}$ & $< 10^{-37}$ \\
\bottomrule
\end{tabular}
\end{sc}
\end{small}
\end{center}
\vspace*{-0.3cm}
\caption{Subjective preference (\%) of 8 raters on 260 audiobook phrases.
Each row reports preferences for a baseline Tacotron vs one of the TP-GST systems.
$t$-test $p$-values are given for both a 3-point and 7-point rating system.\label{table.lessac_sxs_tp_vs_tacotron}}
\end{table}

\vspace*{0.15cm}
\begin{table}[h]
\begin{center}
\begin{small}
\begin{sc}
\setlength\tabcolsep{3.5pt} \begin{tabular}{ccc|cc}
\toprule
\multicolumn{3}{c|}{preference (\%)} & \multicolumn{2}{c} {p-value} \\ 
\hline
TPCW-GST & neutral & TPSE-GST & 3-point & 7-point \\
\midrule
25.1\% &45.4\% &29.4\% &0.063 & 0.054 \\
\bottomrule
\end{tabular}
\end{sc}
\end{small}
\end{center}
\vspace*{-0.3cm}
\caption{Subjective preference (\%) of 8 raters on 260 audiobook phrases for TPCW-GST vs TPSE-GST. $t$-test $p$-values are given for both a 3-point and 7-point rating system.\label{table.lessac_sxs_tpcw_vs_tpse}}
\end{table}
\vspace*{0.25cm}

Table \ref{table.lessac_sxs_tpcw_vs_tpse} shows subjective test results comparing TPCW-GST versus TPSE-GST synthesis. The evaluation sentences, audio clips, and instructions for this test are identical to those used for the previous side-by-side results. These results show that, as expected, raters did not have a strong preference between TPCW-GST and TPSE-GST.

\subsubsection{Automatic Denoising}
\label{sec.expt.singlespeaker.denoising}

About 10\% of the recordings used to train the model for this experiment contain audible high-frequency background noise. This noise is reproduced in many of the 260 baseline Tacotron utterances synthesized for the subjective evaluations above. By contrast, as can be heard on our samples page, both text-prediction pathways completely remove this background noise.

This small empirical finding suggests that TP-GST models can not only separate clean speech from background noise (as demonstrated in \cite{wang2018gst}), but that they can do so without needing a manually-identified ``clean'' style token at inference time. While we did not measure the total number of utterances with and without noise, about 6\% of rater comments mentioned this effect; we also provide a number of examples on our audio samples page.

\subsection{Multiple Speaker Experiments}
\label{sec.expt.multispeaker}

We also present results from a multi-speaker TP-GST system. For these experiments, we use the multi-speaker Tacotron architecture described in \cite{rj2018transfer}, conditioning the model on a $64$-dimensional embedding of the input speaker's identity. 
For training data, we use 190 hours of American English speech, read by 22 different female speakers.
Importantly, the 22 datasets include both expressive and non-expressive speech: to the expressive audiobook data from Section \ref{sec.expt.singlespeaker} (147 hours) we add 21 high-quality proprietary datasets, spoken with neutral prosody. These contain 8.7 hours of long-form news and web articles (20 speakers), and 34.2 hours of of assistant-style speech (one speaker).

The multi-speaker TP-GST uses 40 tokens of dimension 252, and a 6-headed style attention. We train with a minibatch
size of 32 using the Adam optimizer \cite{kingma2014adam}, and perform evaluations at about 250,000 steps.

\begin{figure*}[t]
\centering
\begin{subfigure}{0.95\textwidth}
    \hspace*{0.4cm}
    \includegraphics[scale=0.54]{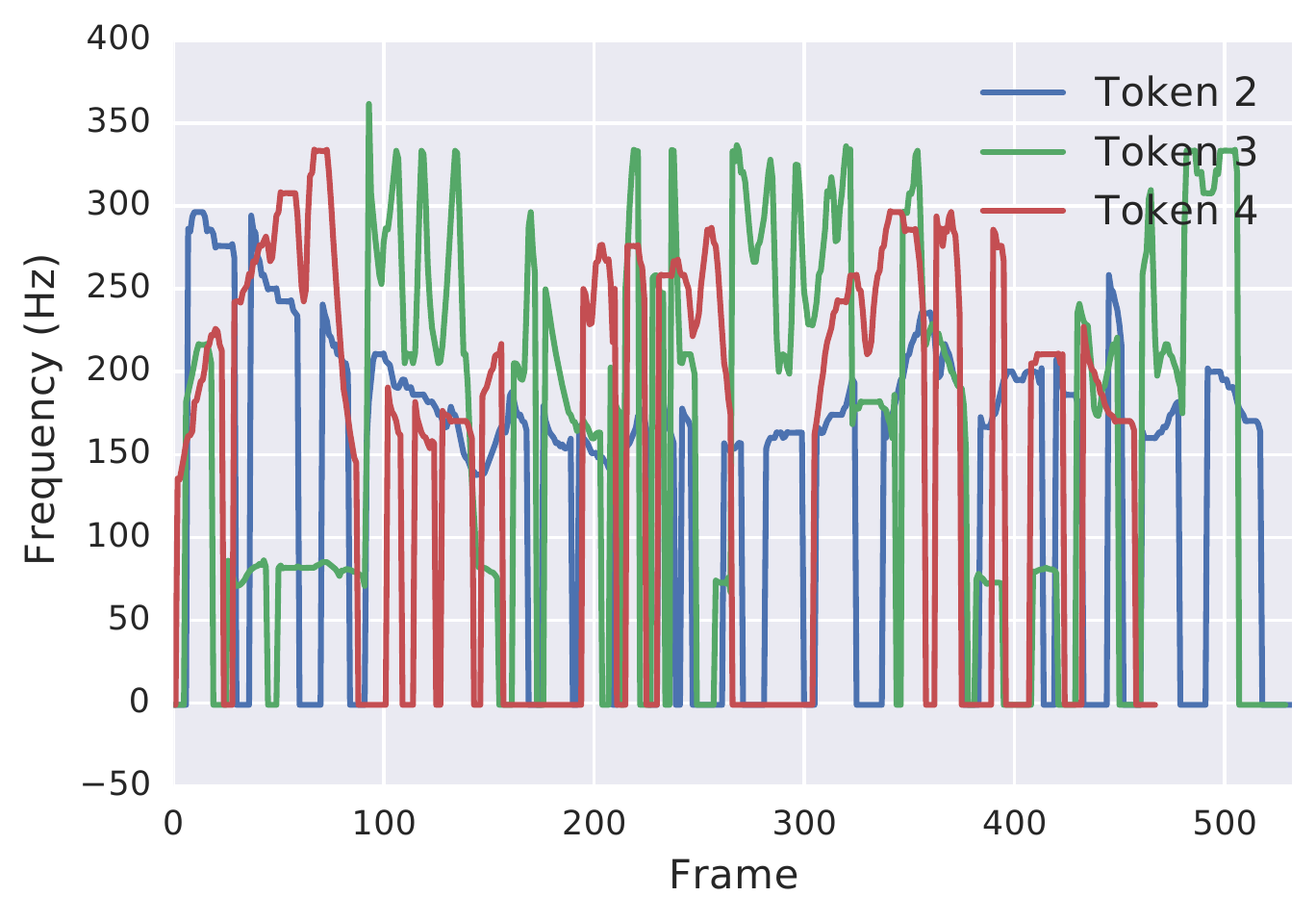}
    \includegraphics[scale=0.54]{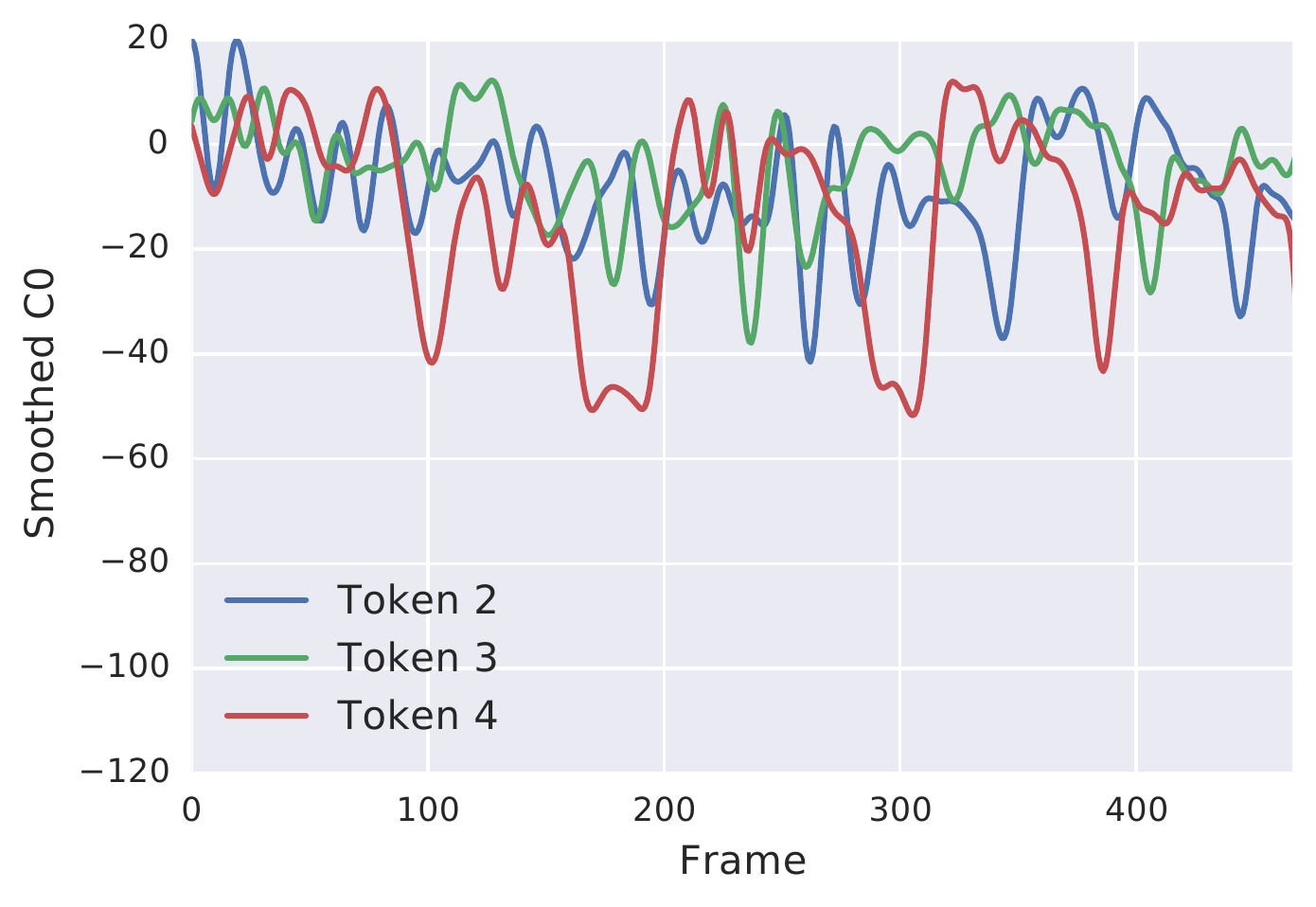}
    \vspace{-0.2cm}
    \caption{$F_0$ and log-$C_0$ for three tokens}
    \vspace{0.3cm}
\end{subfigure}
\begin{subfigure}{1.2\textwidth}    
    \hspace{-0.2cm}
    \includegraphics[scale=0.44]{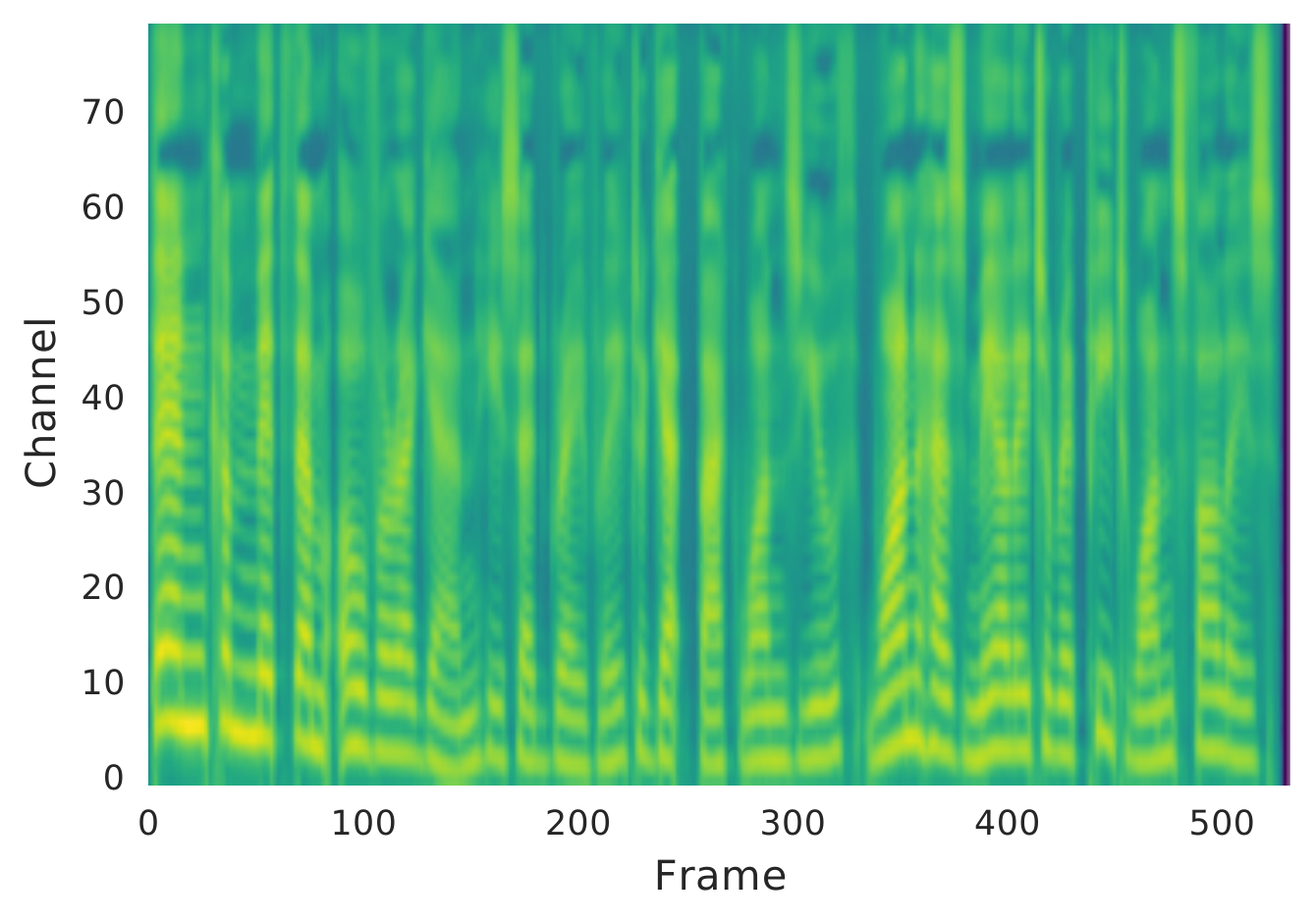}
    \includegraphics[scale=0.44]{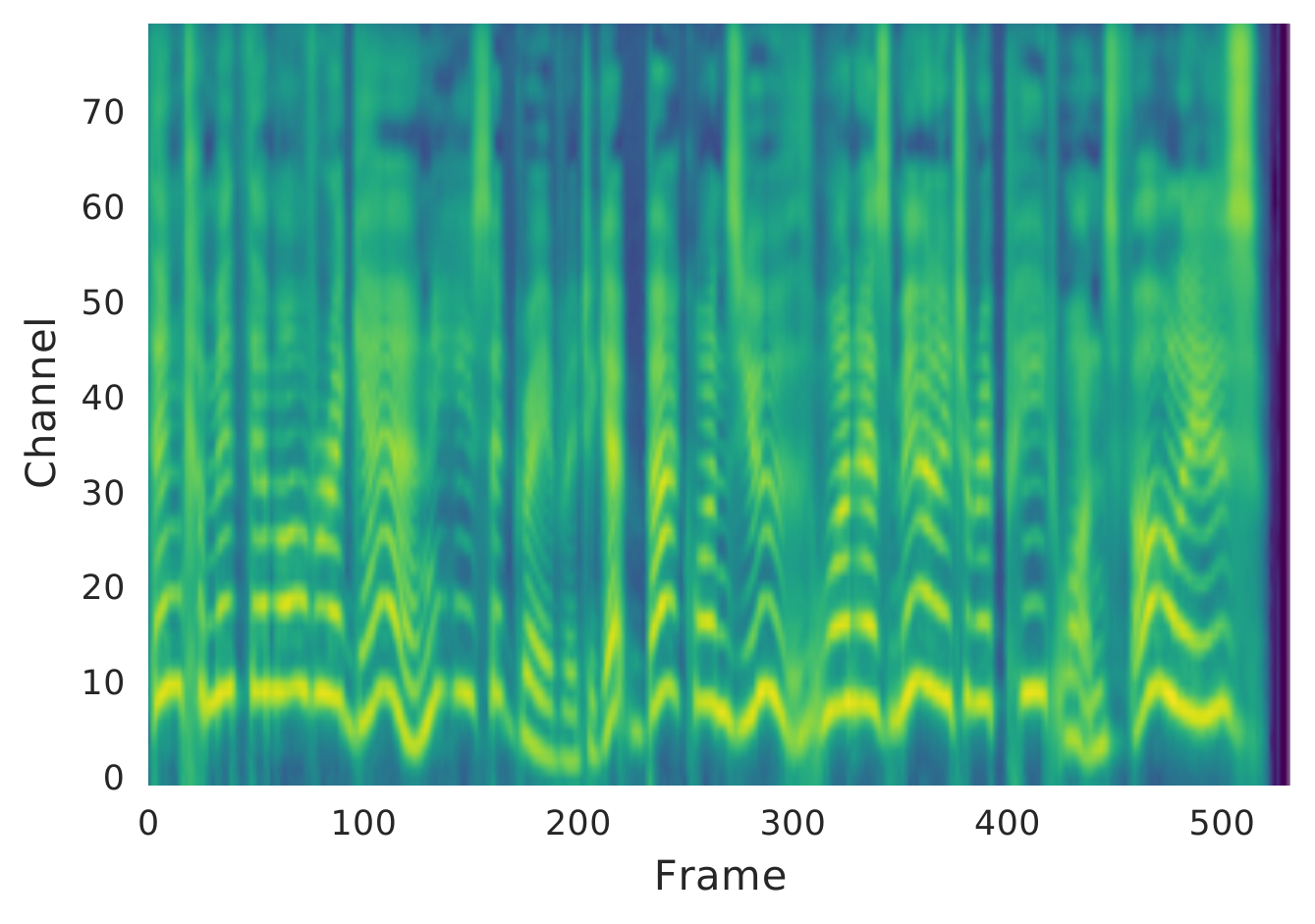}
    \includegraphics[scale=0.44]{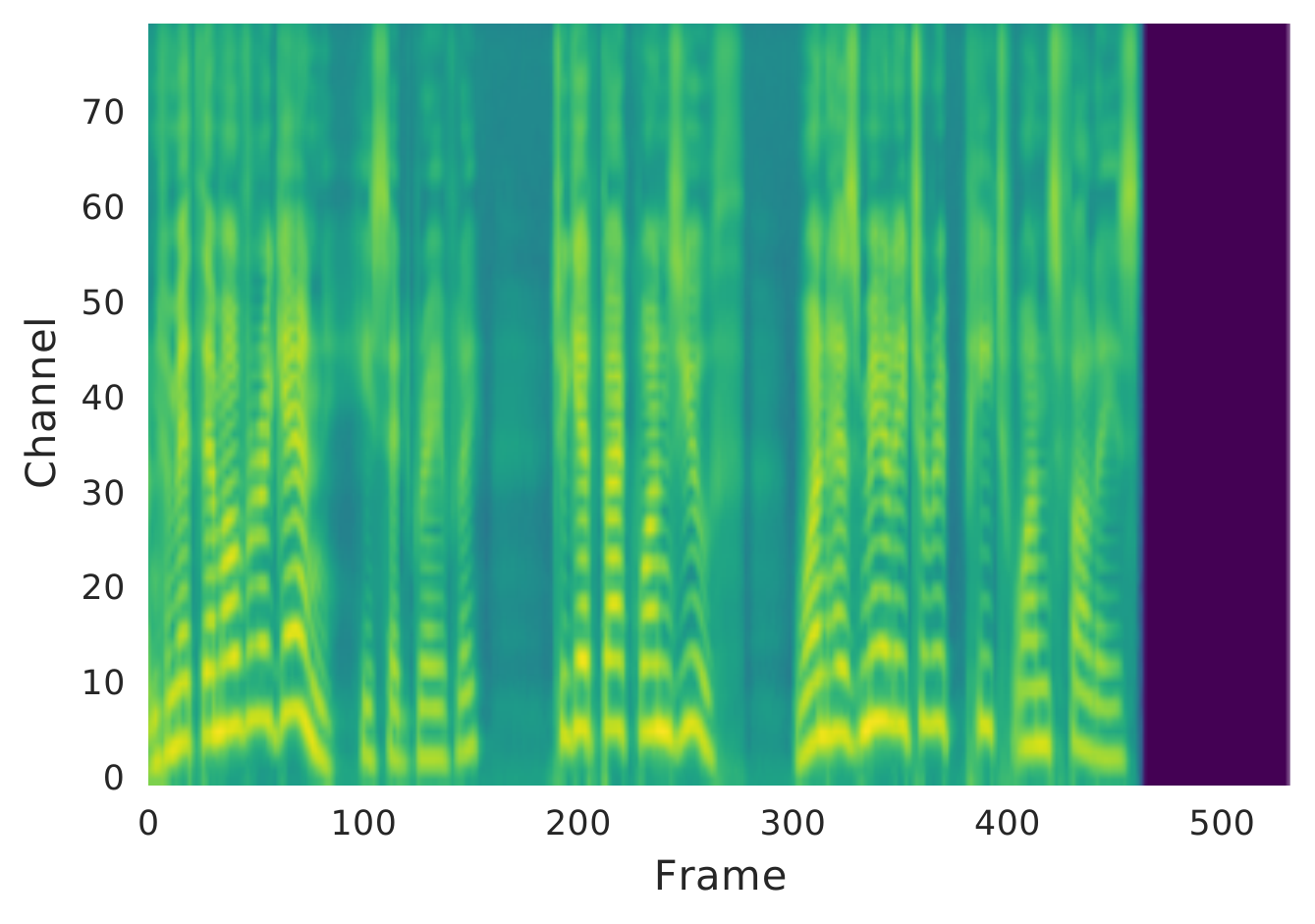}
    \captionsetup{margin={0cm,3.1cm}}
    \caption{Mel spectrograms for the three tokens above}
\end{subfigure}
\caption{(a) $F_0$ and log-$C_0$ of an audiobook phrase, synthesized using three tokens from a multi-speaker text-prediction GST-Tacotron. (b) Mel-scale spectrograms of the same phrase corresponding to each token. See text for details.\label{fig.hol_les_msf_tok40_textpredict_spkr_lessac}}
\par
\par
\end{figure*}

\subsubsection{Shared Multi-Speaker Style Tokens}
Note that, while the multi-speaker TP-GST conditions on speaker identity, style tokens are shared by all speakers. As with the single-speaker models, we can condition on individual tokens at inference time to uncover the factors of variation the model has learned. Figure \ref{fig.hol_les_msf_tok40_textpredict_spkr_lessac} shows $F_0$, log-$C_0$, and spectrograms generated by conditioning on each of three learned style tokens. The model is using the expressive audiobook voice, synthesizing speech from an audiobook unseen during training. Like in the single-speaker case, these plots demonstrate that different tokens capture variation in prosodic factors such as pitch, energy, and speaking rate.  This effect can be heard clearly on our samples page, where individual tokens are synthesized for multiple speakers. These examples show that the learned tokens capture a variety of styles, and, at the same time, that the model preserves speaker identity.  Importantly, conditioning on individual tokens results in style variation even when synthesizing with the neutral-speech voices, despite the fact that these datasets have very little dynamic range for the model to learn.

\subsubsection{Synthesis with Style Prediction}
%Our audio demo page also includes examples of synthesizing phrases from unseen audiobooks from multi-speaker models.
Our audio demo page also includes examples of generating text-predicted style from this multi-speaker TP-GST.
As expected, the Tacotron baseline and TP-GST both generate audio with limited dynamic range when conditioned on the prosodically neutral voice IDs. When synthesizing with the expressive audiobook voice, however, the multi-speaker TP-GST model yields more expressive speech than a multi-speaker Tacotron conditioned on the same data.  While we did not run evaluations comparing these models, we encourage listeners to verify this result for themselves. The samples also reveal that the ``expressive'' multi-speaker TP-GST voice produces similarly expressive speech to that of the single-speaker voice from Section \ref{sec.expt.singlespeaker}, which was trained on the same audiobook data.

\section{Conclusions and Discussion}
\label{sec.conclusion}
In this work, we have shown that a Text-Predicting Global Style Token model can learn to predict speaking style from text alone, requiring no explicit style labels during training, or input signals at inference time. We have demonstrated that TP-GSTs can synthesize audiobook speech in a manner preferred by human raters over baseline Tacotron, 
and that multi-speaker TP-GST models can learn a shared style space while still preserving speaker identity for synthesis.

Future research will explore multi-speaker models more fully, examining how well TP-GSTs can learn factorized representations across genders, accents, and languages. We also plan to investigate larger textual context for prediction, and would like to
learn style representations for both finer-grained and hierarchical temporal resolutions.

Finally, while this work has only investigated style prediction as part of Tacotron, we believe that TP-GSTs can benefit other TTS models, too.  Traditional TTS systems, for example, can use predicted style embeddings as labels, and end-to-end TTS systems can integrate our architecture directly.  More generally, we envision that TP-GSTs can be applied to other conditionally generative models that aim to reconstruct a high-dimensional signal from underspecified input.

\section{Acknowledgements}

The authors thank the Machine Hearing, Google Brain and Google TTS teams for their helpful discussions and feedback.

%\clearpage

%\bibliographystyle{IEEEtran}
%\bibliography{9-references}

\end{document}